\documentclass[10pt,twocolumn,letterpaper]{article}

\usepackage[pagenumbers]{cvpr} % To force page numbers, e.g. for an arXiv version

\usepackage{graphicx}
\usepackage{amsmath}
\usepackage{amssymb}
\usepackage{booktabs}
\usepackage{amsfonts}
\usepackage{bbding} %首先在导言区调用bbding包
\usepackage{color}
\usepackage{multirow}
\usepackage[accsupp]{axessibility}
\usepackage{makecell}
\usepackage[pagebackref,breaklinks,colorlinks]{hyperref}

\usepackage[capitalize]{cleveref}
\crefname{section}{Sec.}{Secs.}
\Crefname{section}{Section}{Sections}
\Crefname{table}{Table}{Tables}
\crefname{table}{Tab.}{Tabs.}

\def\cvprPaperID{13976} % *** Enter the CVPR Paper ID here
\def\confName{CVPR}
\def\confYear{2024}

\begin{document}

\title{EgoVid-5M: A Large-Scale Video-Action Dataset \\ for Egocentric Video Generation}

\author{Xiaofeng Wang~\textsuperscript{\rm 1,2,4}~~~~Kang Zhao~\textsuperscript{\rm 1}~~~~Feng Liu~\textsuperscript{\rm 4}~~~~Jiayu Wang~\textsuperscript{\rm 1}~~~~Guosheng Zhao~\textsuperscript{\rm 2,4}\\Xiaoyi Bao~\textsuperscript{\rm 2,4}~~~~Zheng Zhu~\textsuperscript{\rm 3}~~~~Yingya Zhang~\textsuperscript{\rm 1}~~~~Xingang Wang~\textsuperscript{\rm 2}\\
\textsuperscript{\rm 1}Alibaba 
~ ~ \textsuperscript{\rm 2}CASIA 
~ ~ \textsuperscript{\rm 3}Tsinghua University
~ ~ \textsuperscript{\rm 4}UCAS 
\\
\small{Project Page: \url{https://egovid.github.io}}
\vspace{-2.5em}
}
\twocolumn[{%
\maketitle
\begin{center}
\centering
\resizebox{0.95\linewidth}{!}{
\includegraphics{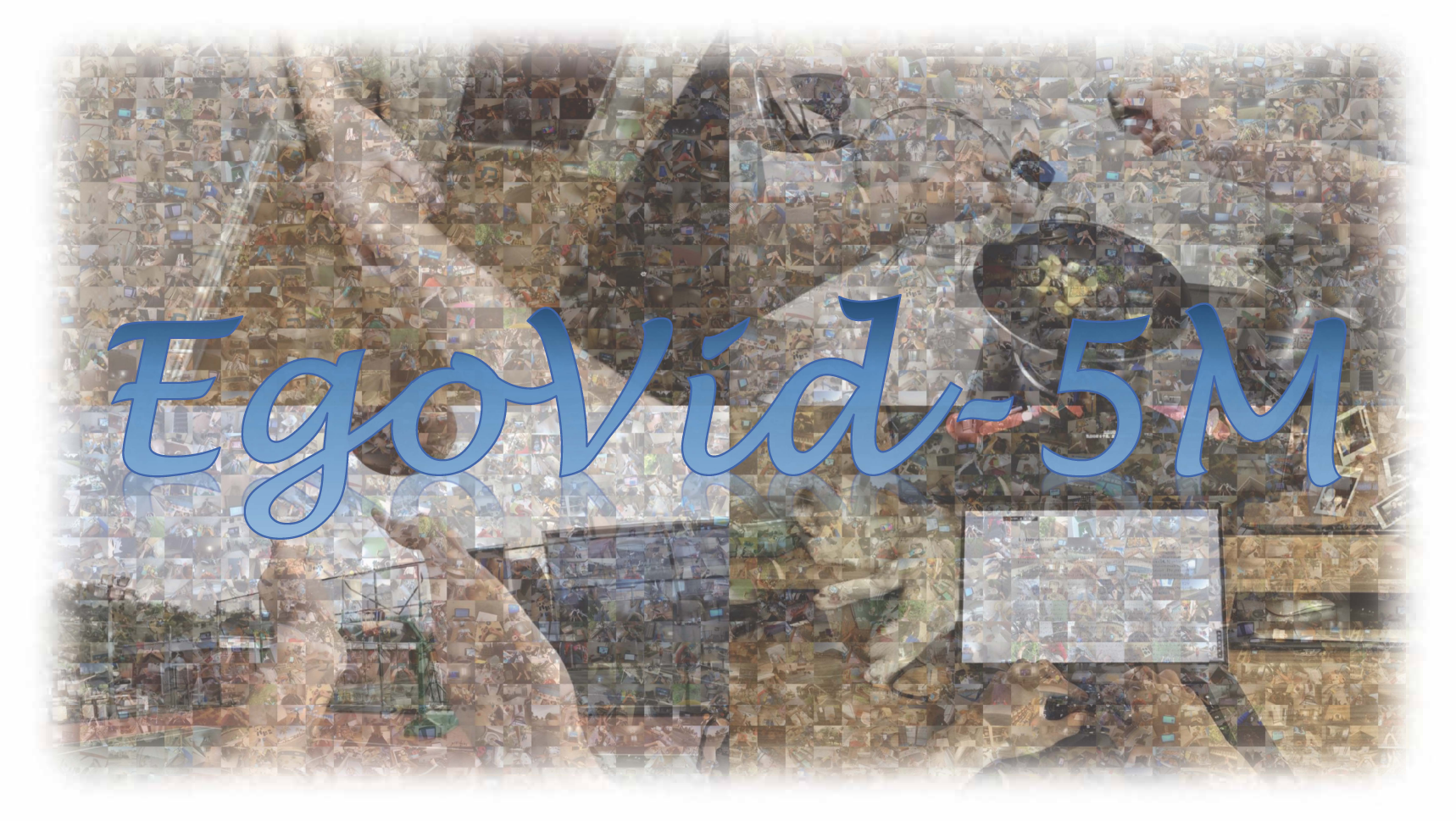}}
\captionof{figure}{\textit{EgoVid-5M} is a meticulously curated high-quality action-video dataset designed specifically for egocentric video generation. It includes detailed action annotations, such as fine-grained kinematic control and high-level textual descriptions. Furthermore, it incorporates robust data cleaning strategies to ensure frame consistency, action coherence, and motion smoothness under egocentric conditions.}
\label{fig:fig0}
\end{center}}]

\maketitle

\begin{table*}[t]
    \centering
    \small
  \resizebox{0.95\linewidth}{!}{
  \begin{tabular}{l|ccccccccc}
    \toprule
    \multicolumn{1}{c|}{Dataset} & Year & Domain & \textit{Gen.}  & Text & Kinematic & \textit{CM.}& \#Videos & \#Frames & Res \\
    \midrule
    HowTo100M~\cite{miech2019howto100m}    & 2019 & Open &\Checkmark & ASR  &\XSolidBrush &\XSolidBrush& 136M   & $\sim 90$  & 240p  \\
   WebVid-10M~\cite{bain2021frozen}        & 2021 & Open &\Checkmark & Alt-Text &\XSolidBrush &\XSolidBrush & 10M  & $\sim 430$ & Diverse\\
    HD-VILA-100M~\cite{xue2022hdvila}     & 2022 & Open  &\Checkmark  & ASR &\XSolidBrush &\XSolidBrush   & 103M   & $\sim 320$  & 720p  \\
    Panda-70M~\cite{chen2024panda}       & 2024 & Open  &\Checkmark & Auto &\XSolidBrush &\XSolidBrush   & 70M  & $\sim 200$  & Diverse \\
    OpenVid-1M~\cite{nan2024openvid}      & 2024 & Open  &\Checkmark  & Auto  &\XSolidBrush &\XSolidBrush& 1M     & $\sim 200$  & Diverse \\
    VIDGEN-1M~\cite{tan2024vidgen1m}       & 2024 & Open  &\Checkmark & Auto &\XSolidBrush &\XSolidBrush & 1M     & $\sim 250$  & 720p \\
    \midrule
    LSMDC~\cite{rohrbach2015lsmdc}         & 2015 & Movie &\XSolidBrush  & Human &\XSolidBrush &\XSolidBrush  & 118K   & $\sim 120$  &  1080p \\
    UCF101~\cite{soomro2012ucf101}         & 2015 & Action  &\XSolidBrush   & Human  &\XSolidBrush &\XSolidBrush & 13K    & $\sim 170$     & 240p \\
    Ego4D~\cite{grauman2022ego4d}          & 2022 & Egocentric  &\XSolidBrush  & Human & IMU &\XSolidBrush  & 931    & $\sim 417K$   & 1080p     \\
    Ego-Exo4D \cite{Grauman_2024_CVPR}     & 2024 & Egocentric &\XSolidBrush & Human & MVS &\XSolidBrush  & 740     & $\sim 186K$   & 1080p     \\
    \textbf{EgoViD-5M (ours)}              & 2024 & Egocentric  &\Checkmark & Auto & VIO &\Checkmark  & 5M   & $\sim 120$   & 1080p     \\
    \bottomrule
    \end{tabular}}
    \caption{\small Comparison of \textit{EgoVid-5M} and other video datasets, where \textit{Gen.} denotes whether the dataset is designed for generative training, \textit{CM.} denotes cleansing metadata,  \#Videos is the number of videos, and \#Frames is the average number of frames in a video.} 
    \label{tab:cmp}
\end{table*}

%%%%%%%%% ABSTRACT
\begin{abstract}
\vspace{-1em}
Video generation has emerged as a promising tool for world simulation, leveraging visual data to replicate real-world environments. Within this context, egocentric video generation, which centers on the human perspective, holds significant potential for enhancing applications in virtual reality, augmented reality, and gaming. However, the generation of egocentric videos presents substantial challenges due to the dynamic nature of egocentric viewpoints, the intricate diversity of actions, and the complex variety of scenes encountered. Existing datasets are inadequate for addressing these challenges effectively.
To bridge this gap, we present \textit{EgoVid-5M}, the first high-quality dataset specifically curated for egocentric video generation. \textit{EgoVid-5M} encompasses 5 million egocentric video clips and is enriched with detailed action annotations, including fine-grained kinematic control and high-level textual descriptions. To ensure the integrity and usability of the dataset, we implement a sophisticated data cleaning pipeline designed to maintain frame consistency, action coherence, and motion smoothness under egocentric conditions.
Furthermore, we introduce \textit{EgoDreamer}, which is capable of generating egocentric videos driven simultaneously by action descriptions and kinematic control signals.
The \textit{EgoVid-5M} dataset, associated action annotations, and all data cleansing metadata will be released 
for the advancement of research in egocentric video generation. \looseness=-1

\end{abstract}

\section{Introduction}

One of the most promising avenues in video generation is the development of world simulators. These systems utilize visual simulations and interactions to deliver applications in the physical world. Contemporary research is increasingly validating the capabilities of video generation in this realm, including applications in autonomous driving \cite{wang2023drivedreamer,hu2023gaia,drivedreamer2,drivewm,drivedreamer4d,yang2024generalized}, autonomous agents \cite{yang2024unisim,zhou2024robodreamer,wu2023daydreamer,hafner2020dreamerV1,hafner2021dreamerV2,hafner2023dreamerv3,bruce2024genie}, and even in general world \cite{videoworldsimulators2024, generalworldmodels}.
In the context of human-centric scenarios, leveraging behavioral actions to drive egocentric video generation has emerged as a pivotal strategy. This approach greatly enhances applications in Virtual Reality (VR), Augmented Reality (AR), and gaming, offering more immersive and interactive experiences and advancing the state of the art in these fields.

Video generation necessitates vast quantities of high-quality data for training. This requirement is even more stringent in egocentric video generation, which is inherently challenging due to the dynamic nature of egocentric perspectives, the richness of observed actions, and the diversity of encountered scenarios. Despite the critical need for specialized data, there is currently a scarcity of publicly available, large-scale datasets for training egocentric video generation models. To bridge this gap, we present the \textit{EgoVid-5M} dataset, a pioneering high-quality dataset specifically curated for egocentric video generation (see Fig.~\ref{fig:fig0}). As shown in Tab.~\ref{tab:cmp}, \textit{EgoVid-5M} is distinguished by several key features: (1) \textbf{High Quality}: This dataset offers 5 million egocentric videos at 1080p resolution. In contrast to Ego4D \cite{grauman2022ego4d}, which is intended for egocentric perception and includes excessive noisy camera motion that is unsuitable for generative training, \textit{EgoVid-5M} undergoes a rigorous data cleaning process. The videos are curated based on stringent criteria, including the alignment between action descriptions and video content, the magnitude of motion, and the consistency between frames. (2) \textbf{Comprehensive Scene Coverage}: \textit{EgoVid-5M} boasts a comprehensive range of scenarios including household environments, outdoor settings, office activities, sports, and skilled operations. It encompasses hundreds of action categories, thus covering the majority of scenes encountered in egocentric perspectives.
(3) \textbf{Detailed and Precise Annotations}: The dataset includes extensive behavioral annotations, which are categorized into fine-grained kinematic control and high-level action descriptions. For kinematic information, we employ Visual Inertial Odometry (VIO) to provide precise annotations, ensuring accurate alignment with video contents. For action descriptions, a multimodal large language model combined with a large language model is utilized to generate detailed text annotations.

\begin{figure*}[t]
\centering
\resizebox{1\linewidth}{!}{
\includegraphics{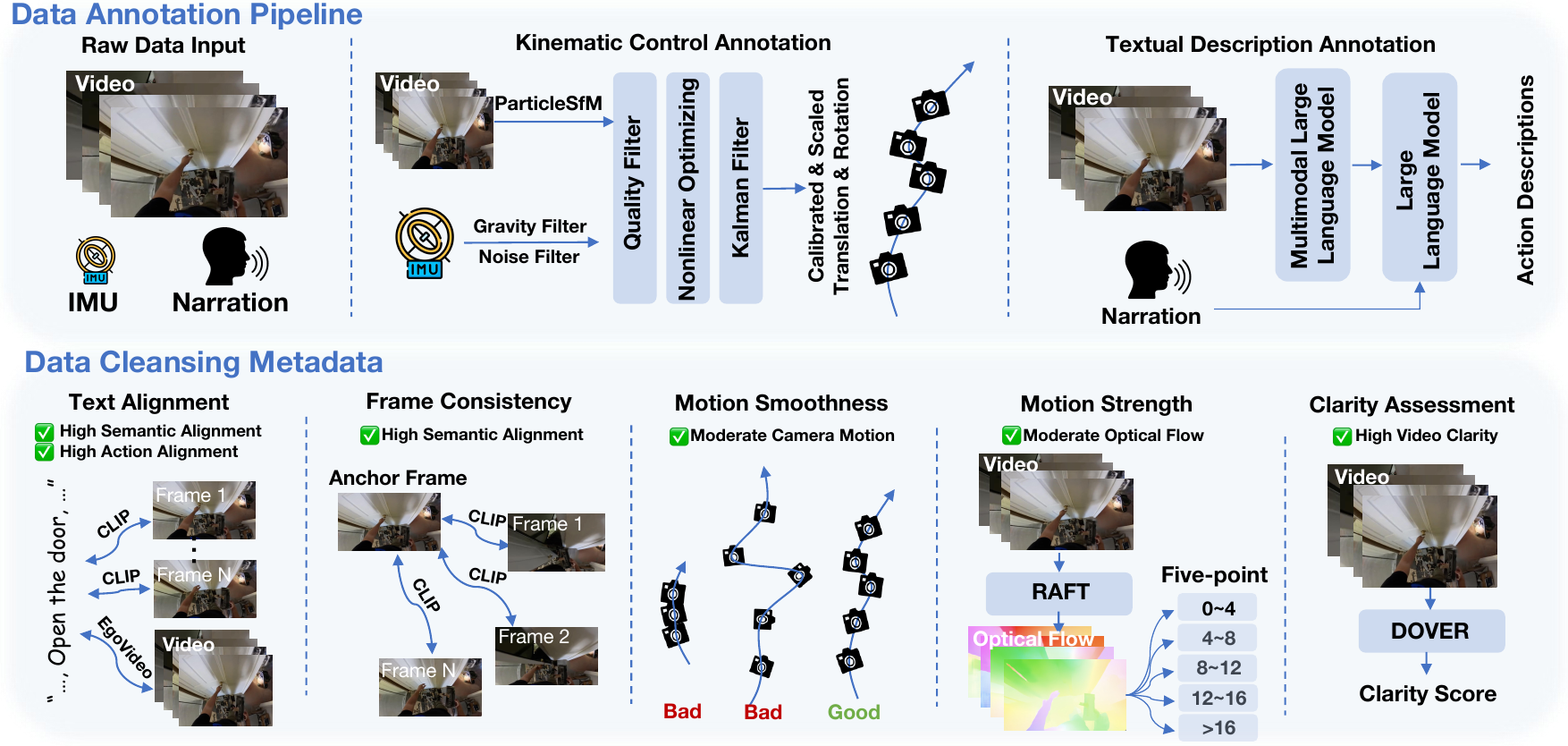}}
\caption{Data annotation pipeline and cleansing metadata of \textit{EgoVid-5M}.}
\label{fig:data}
\vspace{-1em}
\end{figure*}

Leveraging the proposed \textit{EgoVid-5M} dataset, we train different video generation baselines to validate the dataset's quality and efficacy. Various architectures, such as U-Net \cite{svd,xing2023dynamicrafter} and DiT \cite{pku_yuan_lab_and_tuzhan_ai_etc_2024_10948109}, are employed as baseline models, and the experimental results demonstrate that \textit{EgoVid-5M} significantly bolsters the training of egocentric video generation. In addition, we propose \textit{EgoDreamer}, which utilizes both action descriptions and kinematic control to drive the generation of egocentric videos.
To provide a comprehensive assessment of action-driven egocentric video generation, we establish an extensive set of evaluation metrics. These metrics encompass multiple dimensions, including visual quality, frame coherence, semantic compliance with actions, and kinematic accuracy. Extensive experiments show that \textit{EgoVid-5M} markedly enhances the capability of various video generation models to produce high-quality egocentric videos.

The main contributions of this paper can be summarized as follows: (1) We introduce \textit{EgoVid-5M}, the first publicly released, high-quality dataset tailored for egocentric video generation. This dataset is proposed to advance both research and applications in the domain of egocentric visual simulation.
(2) Our dataset includes detailed and precise action annotations, incorporating both fine-grained kinematic control and high-level textual descriptions. In addition, we employ robust data cleaning strategies to ensure frame consistency, action coherence, and motion smoothness within \textit{EgoVid-5M}.
(3) Utilizing \textit{EgoVid-5M}, we conducted extensive experiments on various video generation baselines to validate the dataset's quality and efficacy. Furthermore, to support future advancements in action-driven egocentric video generation, we propose \textit{EgoDreamer}, which leverages both action descriptions and kinematic control to drive egocentric video generation.

\section{Related Work}

\noindent
\subsection{Video Generation as World Simulators}
Video generation technology has seen rapid advancements recently. Both diffusion-based \cite{ho2022imagen,svd,ma2024latte,chen2024videocrafter2,xing2023dynamicrafter,chen2023videocrafter1,opensora,pku_yuan_lab_and_tuzhan_ai_etc_2024_10948109} and token-based \cite{yan2021videogpt,hong2022cogvideo,wang2024worlddreamer,yu2023magvit,yu2023language,sun2024autoregressive} video generation models have proven that the quality and controllability of video generation are steadily improving \cite{zhu2024sora}. Notably, the introduction of the Sora model \cite{videoworldsimulators2024} attracts significant attention which convincingly shows that current video generation models are capable of understanding and adhering to physical laws, thereby substantiating the potential of these models to function as world simulators. This perspective is echoed by Runway, which posits that their Gen-3 Alpha \cite{gen3} is progressing along this promising trajectory. Additionally, video generation models, employed as simulators, have demonstrated significant utility in various real-world applications, including autonomous driving simulations \cite{wang2023drivedreamer,drivedreamer2,drivedreamer4d,drivewm,yang2024generalized,hu2023gaia} and agent-based environments \cite{yang2024unisim,zhou2024robodreamer,wu2023daydreamer,hafner2020dreamerV1,hafner2021dreamerV2,hafner2023dreamerv3,bruce2024genie}. 
Within this context, action-driven egocentric video generation, which centers on the human perspective, holds significant potential for enhancing applications in VR, AR, and gaming. However, current research in the egocentric domain predominantly concentrates on understanding tasks \cite{radevski2023multimodal,plizzari2023can,gong2023mmg,akiva2023self,liu2022hybrid,pei2024egovideo,lin2022egocentric,xu2023egopca,mai2023egoloc}, and generative tasks associated with egocentric scenarios are largely confined to exocentric-to-egocentric video synthesis \cite{li2021ego,luo2024intention,liu2021cross}. This highlights a substantial gap in generating action-driven egocentric videos.
While some methods have explored video generation driven by action interaction \cite{hu2024motionmaster,hou2024training,xu2024camco,he2024cameractrl,wang2024motionctrl,yang2024direct,ma2023trailblazer,hu2023videocontrolnet}, these approaches are mainly concerned with natural scenes featuring smooth camera transitions. This focus limits their ability to model intricate motion patterns inherent in egocentric videos.

\noindent
\subsection{Video Generation Datasets}
In the realm of video generation, the quantity and quality of training data are pivotal for training effective models. Currently, the field of general video generation benefits from several pioneering open-source video datasets. WebVid-10M \cite{bain2021frozen} consists of 52K hours of video, totaling 10.7M text-video pairs. Similarly, InternVid \cite{wang2023internvid} contains over 7M videos spanning nearly 760K hours, resulting in 234M video clips and a comprehensive dataset with 4.1B words in descriptive texts. Panda70M \cite{chen2024panda} stands out with its collection of 70M high-resolution and semantically coherent video samples. OpenVid-1M \cite{nan2024openvid}, offers a million-level, high-quality dataset encompassing diverse scenarios such as portraits, landscapes, cities, metamorphic elements, and animals. In contrast to these general-purpose datasets, specific-scenario datasets typically comprise a limited number of text-video pairs tailored for particular contexts. UCF-101 \cite{soomro2012ucf101} is an action recognition dataset featuring 101 classes and 13,320 total videos. Taichi-HD \cite{siarohin2019first}, a more focused collection, includes 2,668 videos capturing a single person performing Taichi movements.
In the domain of egocentric video generation, existing datasets such as Ego4D \cite{grauman2022ego4d} and Ego-Exo4D \cite{Grauman_2024_CVPR} are primarily designed for egocentric scene understanding tasks and often include excessive noisy camera motion, rendering them unsuitable for generative training. Additionally, EgoGen \cite{li2024egogen}, a synthetic dataset, can not fully encapsulate the complex variations inherent in real-world egocentric views.
To address this gap, we introduce the \textit{EgoVid-5M} dataset, a pioneering and meticulously curated collection designed explicitly for egocentric video generation. \textit{EgoVid-5M} comprises 5M egocentric video clips with precise action annotations and cleansing metadata.

\section{EgoVid-5M}
The training of video generation relies on large-scale, high-quality video data. Therefore, we built \textit{EgoVid-5M} based on the large-scale Ego4D dataset \cite{grauman2022ego4d}. Notably, although Ego4D contains thousands of hours of egocentric videos, it is intended for egocentric perception and includes excessive noisy camera motion that is unsuitable for generative training. Additionally, the narration annotation in Ego4D is overly simplistic and lacks semantic consistency with frames.
To address these issues, we propose a data annotation pipeline that provides detailed and accurate annotations of fine-grained kinematic control and high-level action descriptions. Furthermore, a data cleaning pipeline is developed to ensure alignment between action descriptions and video content, as well as the magnitude of motion and consistency between frames. 

% In the following sections, we will provide a detailed description of the data annotation pipeline and the data cleansing pipeline.

\subsection{Data Annotation Pipeline}
\label{sec:ann}
In order to simulate egocentric videos from actions, we construct detailed and accurate action annotations for each video segment, encompassing low-level kinematic control (e.g., ego-view translation and rotation), as well as high-level textual descriptions. The annotation pipeline is shown in the upper part of Fig.~\ref{fig:data}.

\noindent
\textbf{Kinematic Control Annotation}
In order to accurately describe complex egocentric movements, we utilize the Visual-Inertial Odometry (VIO) method to construct kinematic control signals. This involves using ParticleSfM \cite{zhao2022particlesfm} to obtain scale-ambiguous camera poses $P_c$ from video, followed by integrating IMU signals $\{{I_t}\}_{t=0}^{T-1}$ to obtain more accurate and scaled camera poses. However, there are several challenges to overcome. (1) The IMU signals are subject to noise. (2) The transformation matrix between the IMU and the camera is unknown. (3) The initial velocity of the IMU is unknown. (4) The scale factor of the $P_c$ is unknown.
To address the aforementioned problems, we first utilize high-pass Butterworth filters $\mathcal{F}_{IFFT}(\mathcal{H}_\text{low}(s) \cdot \mathcal{F}(s))$ and low-pass Butterworth filters $\mathcal{F}_{IFFT}(\mathcal{H}_\text{high}(s) \cdot \mathcal{F}(s))$ to filter out the gravity signal and high-frequency noise,
% \begin{equation}
%     \mathcal{F}_{IFFT}(\mathcal{H}_\text{low}(s) \cdot \mathcal{F}(s)),
% \end{equation}
% \begin{equation}
%     \mathcal{F}_{IFFT}(\mathcal{H}_\text{high}(s) \cdot \mathcal{F}(s)),
% \end{equation}
where $\mathcal{F}(s)=\mathcal{F}_{FFT}(I)$ is the \textit{Fast Fourier Transform} and $\mathcal{F}_{IFFT}$ is the inverse operation. $\mathcal{H}_\text{low}(s)=\frac{1}{1+(\frac{s}{w_c})^{2n}}$ is the low-pass filter, $\mathcal{H}_\text{high}(s)=\frac{(\frac{s}{w_c})^{2n}}{1+(\frac{s}{w_c})^{2n}}$ is the high-pass filter, $w_c$ represents the cutoff frequency while $n$ represents the filter order. Next, we propose a quality filter to drop the low-quality $P_c$ and $I$, where the motivation is that the number of reconstructed points $N_p$ (generated from ParticleSfM) is a reflection of the accuracy of $P_c$ \cite{xu2024camco}, and the variance of IMU reflects the dynamic nature of the video. Therefore, the retained data needs to simultaneously satisfy $N_p \ge N_\text{thres}$ and $\frac{1}{T}\sum_{t=0}^{T-1}(I_t-\overline{I})^2 \le V_\text{thres}$. Next, we perform the least squares minimization with $P_c$ and the integrated IMU signal $\{{I_t}\}_{t=0}^{T-1}$ to calculate the initial velocity $v(0)$ of the IMU signal, the transformation matrix $T_{I}$ from IMU to the camera, and the scale factor $\lambda$ of the $P_c$:
\begin{equation}
    \min_{v_0,T_{I},\lambda}|T_{I}P_I(T-1)-\lambda P_c|^2,
\end{equation}
where $P_I(T-1)$ can be derived from:
\begin{align}
P_I(t+1) & = P_I(t) + v(t)\Delta t + \frac{1}{2}I(t)\Delta t^2, \\
v(t+1) & = v(t) + I(t) \Delta t,
\end{align}
with the initial condition $P(0)=\mathbf{0}$.
Finally, we utilize the Kalman filter to fuse these two signals under the camera coordinate (see supplement for more details).

\begin{figure*}[t]
\centering
\resizebox{1\linewidth}{!}{
\includegraphics{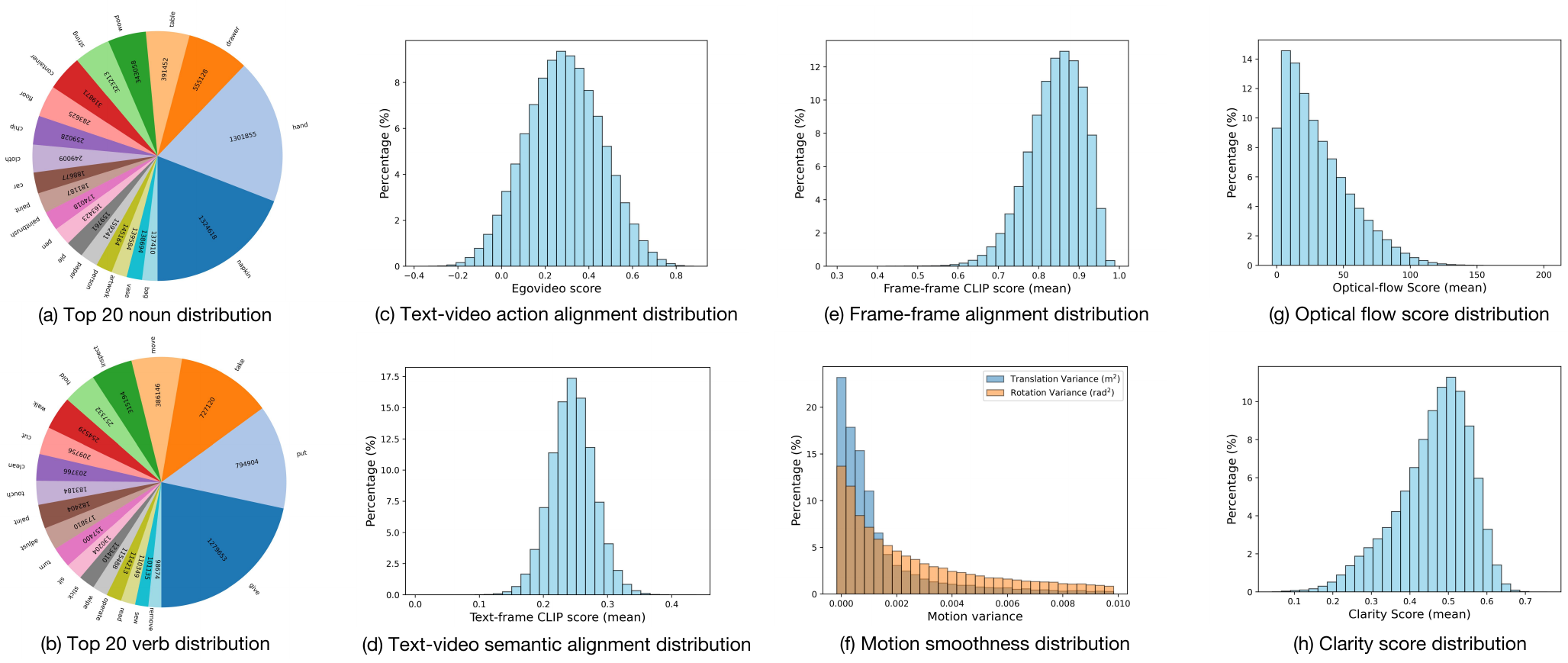}}
\caption{Data annotation distribution of \textit{EgoVid-5M}. (a) and (b) describe the quantities of the top 20 verbs and nouns. (c) Text-video action alignment is assessed using the EgoVideo score. (d) and (e) measure the semantic similarity between text and frames and between frames and the first frame using the average CLIP score.
(f) Motion smoothness is quantified by the variance of translation and rotation.
(g) Motion strength is represented by the average global optical flow.
(h) Video clarity is determined by the DOVER score.}
\label{fig:dist}
\vspace{-1em}
\end{figure*}

\noindent
\textbf{Textual Description Annotation}
In addition to kinematic control, another supplementary information of egocentric action is textual descriptions. In the Ego4D dataset, only human narrations serve as text annotations, but the narrations are relatively simple and lack semantic consistency with frames (see supplement). Therefore, we utilize a multimodal large language model (MLLM) to provide detailed action captions for the videos. Considering that existing open-source multimodal language models are not as proficient in following instructions as large language models (LLM), we first prompt LLaVA-NeXT-Video-32B-Qwen \cite{zhang2024llavanext-video} to provide detailed captions for videos (including foreground, background, main subjects, and action information). Then, we prompt Qwen2 \cite{qwen2} to summarize egocentric action descriptions from the aforementioned captions, with human narrations as the supplementary prompt. Through the combination of MLLM and LLM, our textual descriptions can accurately describe egocentric action while ensuring semantic consistency.
We also utilize LLM to analyze the \textit{Nouns} and \textit{Verbs} in each textual description, and classify them into hundreds of action categories (as shown in the Fig.~\ref{fig:dist}(a)-(b)). The resulting textual descriptions include actions in household environments, outdoor settings, office activities, sports, and skilled operations, thus covering the majority of scenes encountered in egocentric perspectives.\looseness=-1

\subsection{Data Cleaning Pipeline}
\label{sec:clean}
The data quality significantly influences the effectiveness of training generative models. Prior works \cite{nan2024openvid,svd,tan2024vidgen1m} have delved into various cleaning strategies to improve video datasets, focusing on aesthetics, semantic coherence, and optical flow magnitude. Based on these cleaning strategies, this paper presents a specialized cleaning pipeline specifically designed for egocentric scenarios. The pipeline is illustrated in the lower part of Fig.~\ref{fig:data}.

\begin{figure*}[t]
\centering
\resizebox{0.9\linewidth}{!}{
\includegraphics{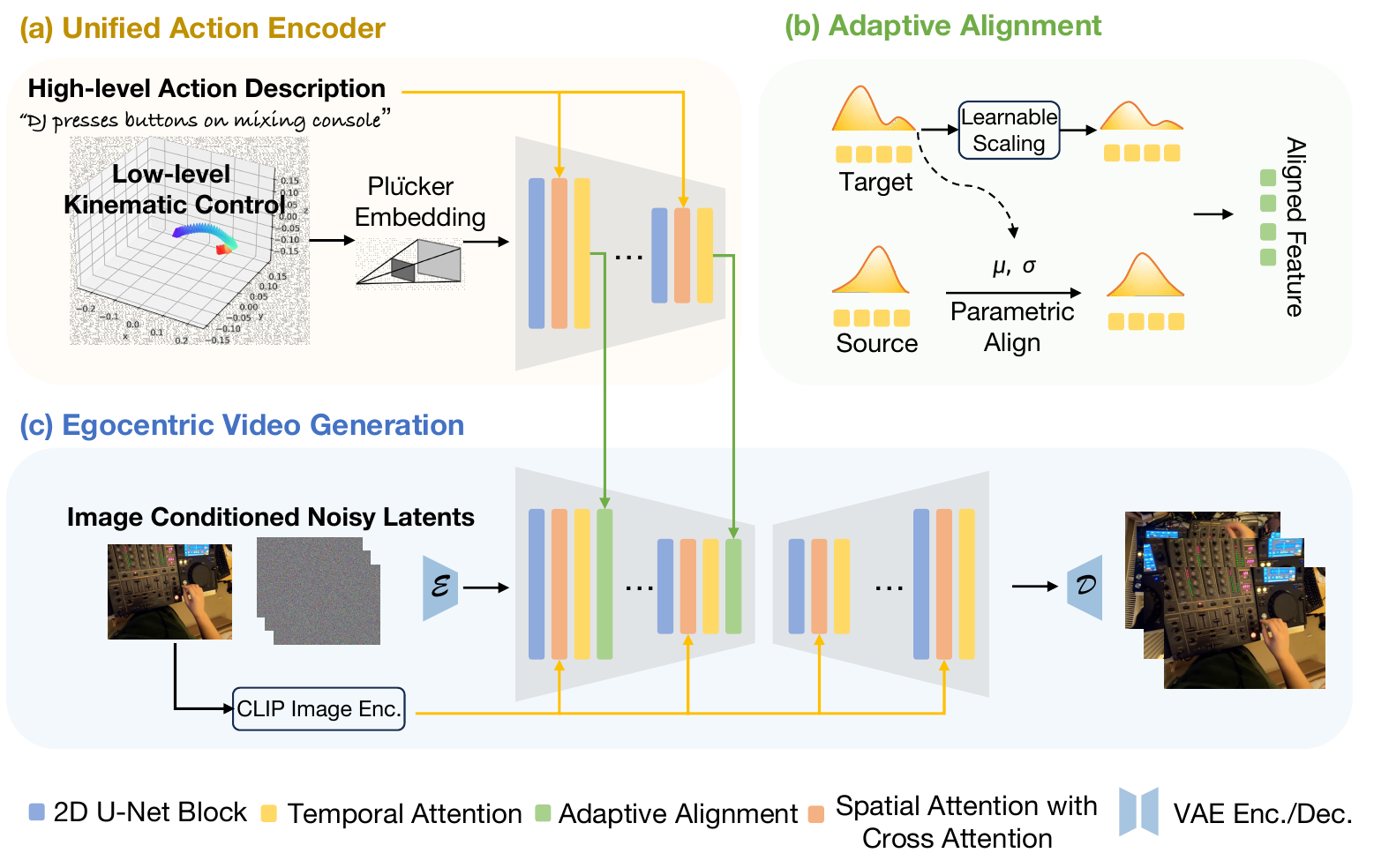}}
\caption{The overall framework of \textit{EgoDreamer}. \textit{EgoDreamer} introduces (a) the Unified Action Encoder to embed different action inputs simultaneously, and it utilizes (b) the Adaptive Alignment to integrate action conditions into the egocentric video generation branch (c).}
\label{fig:egodreamer}
\vspace{-1.5em}
\end{figure*}

\noindent
\textbf{Text-video Consistency}
We utilize CLIP EgoVideo \cite{pei2024egovideo} and \cite{radford2021learning} to evaluate the alignment between textual descriptions and video frames, leveraging EgoVideo's focus on action alignment and CLIP's emphasis on global semantic similarity. In particular, evenly-spaced frames are gathered to calculate the EgoVideo similarity with the text. (refer to Fig.~\ref{fig:dist}(c) for Egovideo score distribution). Subsequently, these four frames are separately extracted to calculate CLIP similarity with the corresponding text (see Fig.~\ref{fig:dist}(d) for CLIP similarity score distribution).

\noindent
\textbf{Frame-frame Consistency}
The higher the semantic consistency between video frames, the more conducive it is to generative training. To analyze this relationship, we uniformly extract three frames alongside the first frame to compute frame CLIP similarity. The distribution of semantic consistency is illustrated in Fig.~\ref{fig:dist}(e).

\noindent
\textbf{Motion Smoothness}
Excessive egocentric motion can lead to video fluctuations, which is detrimental to training visual generation models. To address this issue, we propose measuring the degree of translation variation $\frac{1}{T}\sum_{t=0}^{T-1}(Tr_t-\overline{Tr})^2$ and rotation variation $\frac{1}{T}\sum_{t=0}^{T-1}(Ro_t-\overline{Ro})^2$ to quantify motion smoothness, where $Tr$ and $Ro$ are translation and rotation measured in Sec.~\ref{sec:ann} (see motion smoothness distribution in Fig.~\ref{fig:dist}(f)).

\noindent
\textbf{Motion Strength}
A typical approach to describe video motion strength is optical flow \cite{teed2020raft}. Therefore, we first represent video motion by averaging global optical flow (see motion strength distribution in Fig.~\ref{fig:dist}(g)), we additionally calculate the \textit{five-point} optical flow, which includes the proportion of optical flow score across pixel intervals: 0–4, 4–8, 8–12, 12–16, and above 16 (more details see supplement). This method offers a multi-faceted perspective on motion strength, addressing both the movement of small foreground objects and the overall camera motion.

\noindent
\textbf{Clarity Assessment}
For egocentric scenes, clarity and realism are paramount. Therefore, instead of relying on CLIP for aesthetic scoring \cite{clip-aesthetic}, we apply DOVER \cite{wu2023dover} to assess video clarity (refer to Fig.~\ref{fig:dist} for DOVER score distribution), prioritizing visual sharpness and detail in our dataset.

Based on the cleansing metadata, we vary thresholds to filter and obtain high-quality training data. Specifically, experiments are conducted in Sec.~\ref{sec:exp_filter} to explore the effects of three mainstream cleaning strategies on egocentric video generation training. Additionally, given the significance of data cleaning strategies in training video generation models \cite{nan2024openvid,svd,tan2024vidgen1m}, and the substantial computational cost—\textbf{thousands of GPU days}—to annotate and clean millions of videos, we release all annotation and cleansing metadata to encourage community research into the impact of various cleaning strategies on egocentric video training.

\begin{figure*}[t]
\centering
\resizebox{1\linewidth}{!}{
\includegraphics{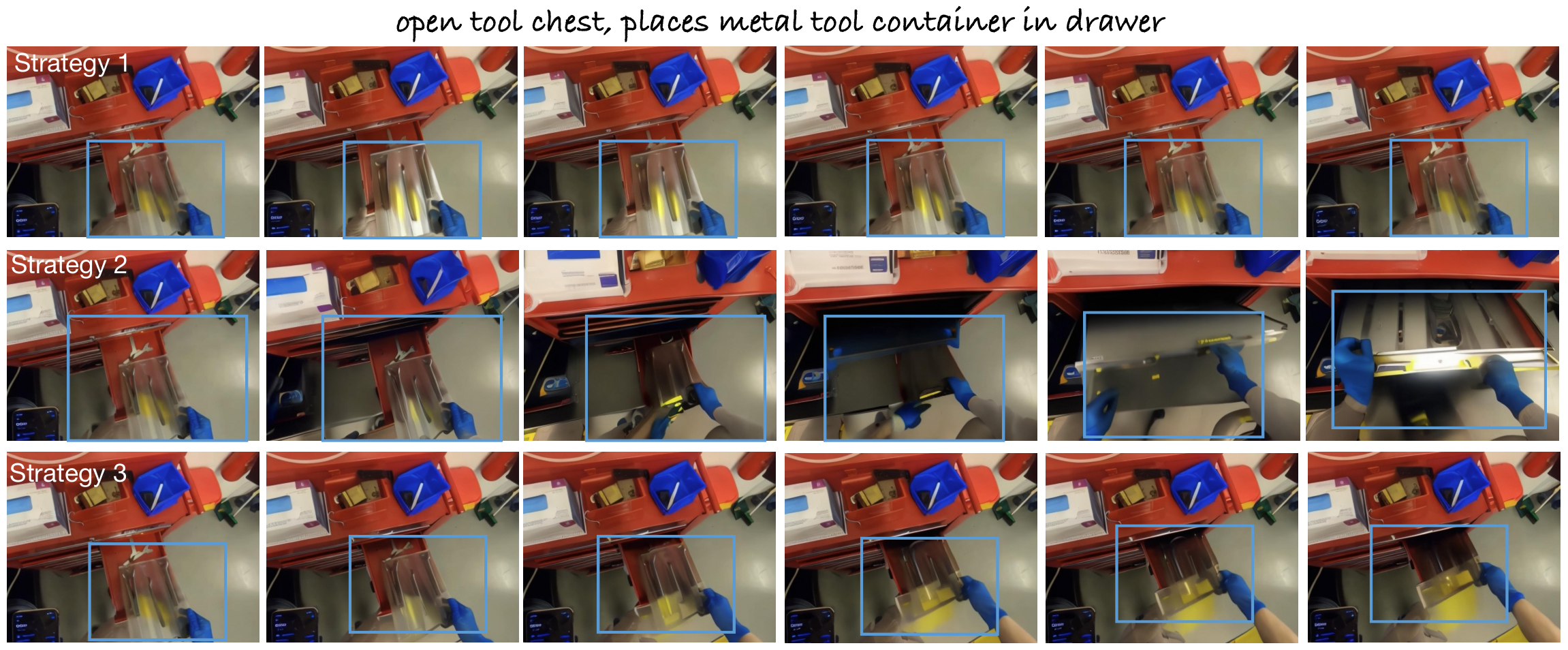}}
\caption{The video visualization comparison across different data cleaning strategies reveals distinct outcomes, where the \textcolor{blue}{blue} box highlights the difference. Videos generated by strategy-1 fail to capture local motion and tend to be stationary. In contrast, videos produced by strategy-2 exhibit excessive motion, compromising semantic coherence. Meanwhile, videos generated by strategy-3 effectively model intricate hand movements, striking a balance between motion strength and semantic fidelity.}
\label{fig:clean_cmp}
\end{figure*}

\begin{figure}[t]
\centering
\resizebox{1\linewidth}{!}{

\includegraphics{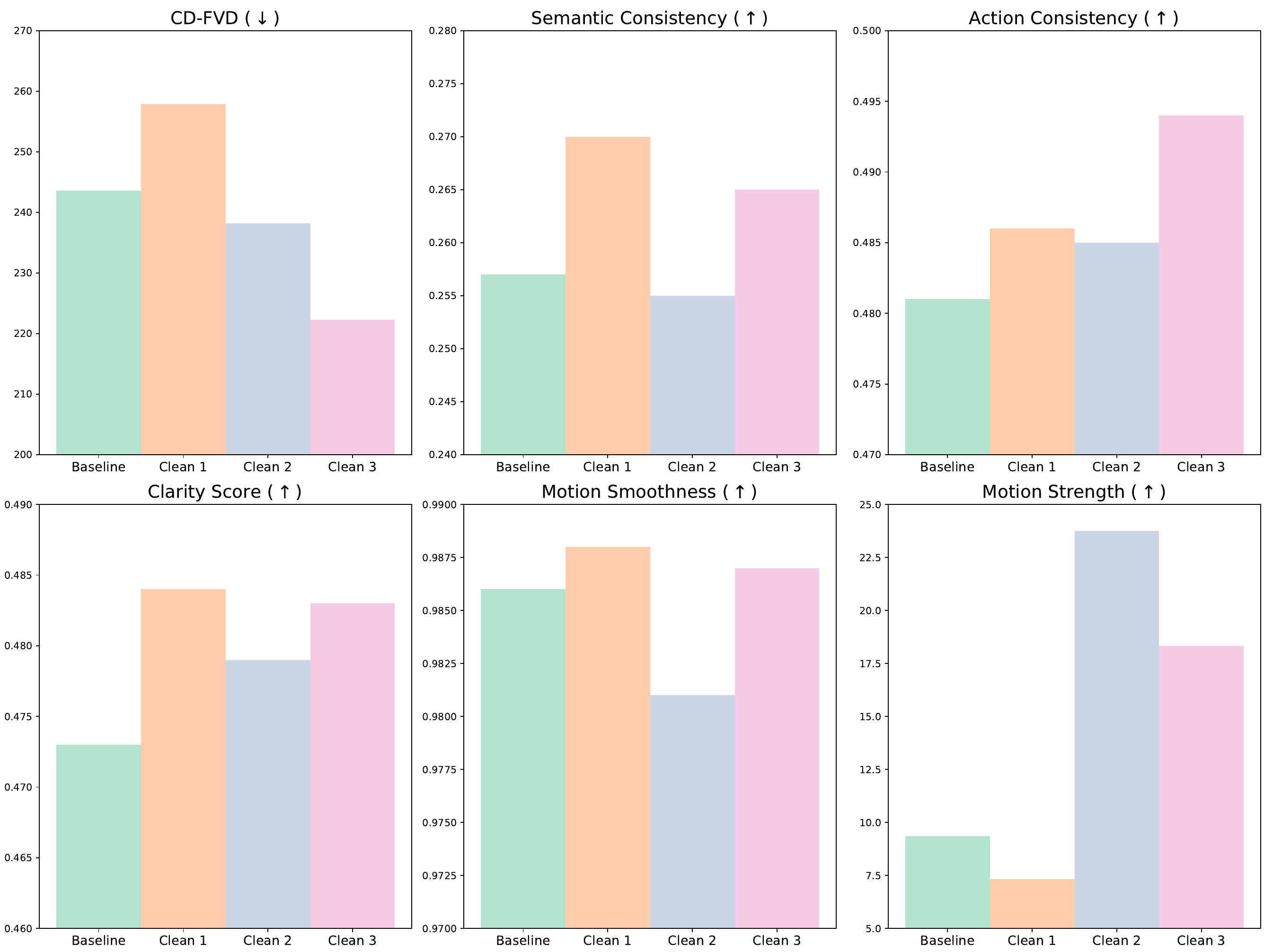}}
\caption{Video generation quantitative comparisons between different data cleaning strategies, where the baseline is DynamiCrafter \cite{xing2023dynamicrafter} initialized with its original weights.}
\label{fig:clean}
\vspace{-1em}
\end{figure}

\section{EgoDreamer}
In the context of ego-centric world simulators, action-driven video generation is paramount. However, existing action-driven video generation approaches \cite{hu2024motionmaster,hou2024training,xu2024camco,he2024cameractrl,wang2024motionctrl,yang2024direct,ma2023trailblazer,hu2023videocontrolnet} primarily focus on camera movements within static scenes, making it challenging to model complex ego-motion. Therefore, we propose \textit{EgoDreamer}, which can produce egocentric videos driven simultaneously by high-level action descriptions and low-level kinematic control. As illustrated in Fig.~\ref{fig:egodreamer}, \textit{EgoDreamer} adopts a similar architecture of \cite{xing2023dynamicrafter} to enable image-conditioned video generation. Besides, \textit{EgoDreamer} features two key innovations: (1) It introduces a Unified Action Encoder (UAE) that embeds two distinct action inputs, allowing for a more nuanced representation of ego movements. (2) It leverages Adaptive Alignment (AA) that encapsulates multi-scale control signals in the parametric alignment perspective, enhancing the action control efficacy.

\noindent
\textbf{Unified Action Encoder.} In this framework, the UAE simultaneously encodes both low-level and high-level actions. Specifically, it first utilizes $\text { Plücker }$ embedding \cite{he2024cameractrl,sitzmann2021light} to encode kinematic signals:
\begin{align}
    \mathbf{p}_{u, v}&=\left(\mathbf{t} \times \mathbf{d}_{u, v}, \mathbf{d}_{u, v}\right), \\
    \mathbf{d}_{u, v}&=\mathbf{R K}^{-\mathbf{1}}[u, v, 1]^{T}+\mathbf{t},
\end{align}
where $\mathbf{R}$ and $\mathbf{t}$ is the rotation matrix and translation vector, $\mathbf{K}$ is the intrinsic matrix, and $\mathbf{p}_{u, v}$ is the $\text { Plücker }$ embedding at pixel $(u, v)$. Then, low-level signal $\mathbf{p}$ is encoded through a series of U-Net blocks, while a high-level action description $d$ is simultaneously embedded via CLIP \cite{radford2021learning} and cross-attention mechanisms. The action output $A$ of one U-Net block can be formulated as:
\begin{equation}
    A=\mathcal{F}_t(\mathcal{F}_c(\mathcal{F}_s(\mathcal{F}_\text{conv}(\mathbf{p})), \text{CLIP}(d))),
\end{equation}
where $\mathcal{F}_t$ is the temporal self-attention, $\mathcal{F}_c$ is the cross-attention, $\mathcal{F}_s$ is the spatial self-attention, $\mathcal{F}_\text{conv}$ is the 2D convolution block. Notably, previous methods \cite{xu2024camco,he2024cameractrl} encode text and kinematics separately, ignoring that low-level kinematics and high-level action descriptions are coupled.
In contrast, the proposed UAE focuses on modeling the relationship between different action inputs, thus the generated action control signals capture both camera movements and complex egocentric dynamics  (e.g., hand interactions).

\noindent
\textbf{Adaptive Alignment.}
Based on the multi-scale U-Net architecture, the UAE outputs multi-scale $\{A_i\}_{i=0}^{3}$. Then \textit{EgoDreamer} encapsulates control signals in the perspective of parametric alignment:
\begin{equation}
    L_i = \alpha L_i + \frac{A_i-\mu_L}{\sigma_L},
\end{equation}
where $L_i$ is the output of one U-Net block in the main Diffusion branch, $\alpha$ is a learnable parameter, $\mu_L,\sigma_L$ are the mean and standard deviation of $L_i$. The introduced AA module is inspired by cross normalization \cite{peng2024controlnext} and applies it to multi-scale U-Net feature alignment. Compared to ControlNet's zero-initialization \cite{zhang2023adding}, our method achieves better control effectiveness.

\begin{figure*}[t]
\centering
\resizebox{0.95\linewidth}{!}{
\includegraphics{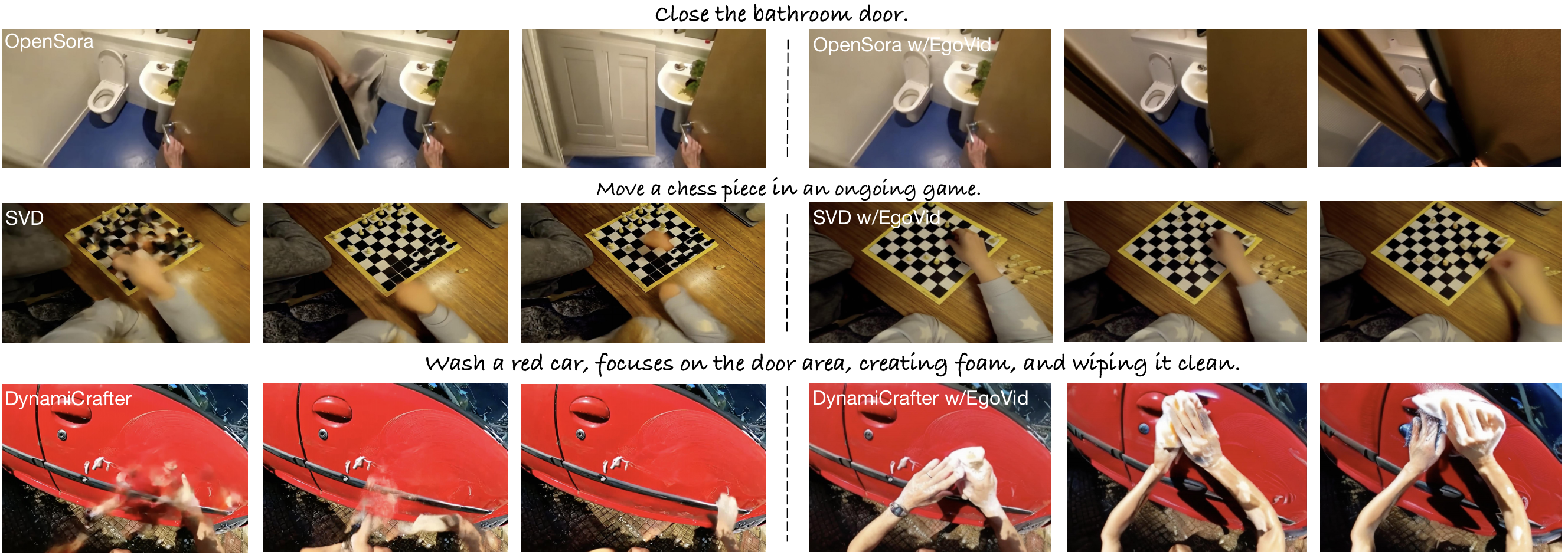}}
\caption{Visualizations demonstrate that \textit{EgoVid}-fintuned baselines (OpenSora \cite{opensora}, SVD \cite{svd}, DynamiCrafter \cite{xing2023dynamicrafter}) generate egocentric videos with stronger frame-consistency and better semantic-alignment.}
\vspace{-0.5em}
\label{fig:vis_cmp}
\end{figure*}
\begin{table*}[t]
    \centering
    \small
  \resizebox{1\linewidth}{!}{
  \begin{tabular}{l|c|cccccc}
    \toprule
    \multicolumn{1}{l|}{Method} & w. \textit{EgoVid} & CD-FVD $\downarrow$ & Semantic Consistency $\uparrow$& Action Consistency $\uparrow$  & Clarity Score $\uparrow$ & Motion Smoothness $\uparrow$ & Motion Strength $\uparrow$\\
    \midrule
    SVD~\cite{svd} & \XSolidBrush & 591.61 & 0.258 & 0.465 & 0.479 & 0.971 & 18.897\\
    SVD~\cite{svd} & \Checkmark & \textbf{548.32} & \textbf{0.266} & \textbf{0.471} & \textbf{0.485} & \textbf{0.974} & \textbf{21.032} \\
    \midrule
    DynamiCrafter~\cite{xing2023dynamicrafter} & \XSolidBrush & 243.63 &  0.257 & 0.481 & 0.473 & 0.986 & 9.357\\
    % DynamiCrafter~\cite{xing2023dynamicrafter} & \Checkmark & 249.46/236.82 & 0.256/0.255 &  0.487/0.494 & 0.439/0.445 & 0.982/0.980 & 21.426/22.149\\
    DynamiCrafter~\cite{xing2023dynamicrafter} & \Checkmark & \textbf{236.82} & \textbf{0.265} & \textbf{0.494} & \textbf{0.483} & \textbf{0.987} & \textbf{18.329}\\
    \midrule
    OpenSora~\cite{opensora} & \XSolidBrush & 809.46 & 0.260 & 0.489 & 0.520 &0.983 & 7.608 \\
    OpenSora~\cite{opensora} & \Checkmark & \textbf{718.32} & \textbf{0.266} & \textbf{0.494} & \textbf{0.528} & \textbf{0.986} & \textbf{15.871} \\
    \bottomrule
    \end{tabular}}
    \caption{\textit{EgoVid} significantly enhances egocentric video generation. Experimental results demonstrate that training with \textit{EgoVid} improves performance across all three baselines on six metrics.} 
    \vspace{-1em}
    
    \label{tab:exp_cmp}
\end{table*}

\section{Experiment}
\subsection{Experiment Details}

\noindent
\textbf{Dataset.} The proposed \textit{EgoVid-5M} dataset is partitioned as the training set and the validation set. For the validation set, we select samples with high text-video semantic consistency, moderate video motion, high video clarity, and diverse scene coverage including household environments, outdoor settings, office activities, sports, and skilled operations. This resulted in a final validation set \textit{EgoVid-val} with 1.2K samples, with a training set \textit{EgoVid-train} with 4.9M samples. Notably, due to the known issue in Ego4D IMU annotation\footnote{https://ego4d-data.org/docs/data/imu/}, we annotate kinematic controls for 65K video samples with accurate IMU data. The annotated subset \textit{EgoVid-65K} is $\sim5\times$ larger than the current largest kinematic annotation dataset \cite{xu2024camco}, which is utilized further to train the ability of kinematic control video generation.

\noindent
\textbf{Training}
We validate the effectiveness of our \textit{EgoVid-5M} using video diffusion baselines with different architectures, including U-Net (SVD \cite{svd} and DynamiCrafter \cite{xing2023dynamicrafter}), and DiT (OpenSora \cite{opensora}). Building upon these pre-trained models, we employe a continuous training approach to train 480p videos for enhanced training efficiency. For \textit{EgoDreamer}, we first initialize it with pre-trained weights \cite{xing2023dynamicrafter},
then \textit{EgoDreamer} are further trained on \textit{EgoVid} to adapt to egocentric scenes. Finally, we finetune the proposed UAE and AA using \textit{EgoVid-65K}. All experiments are conducted on NVIDIA A800 GPUs. For additional training details, please refer to the supplementary materials.

\noindent 
\textbf{Evaluation.} We adopt a set of metrics from AIGCBench \cite{fan2023aigcbench} and VBench \cite{huang2024vbench} to assess the quality of the generated egocentric videos. Specifically, our evaluation metrics utilize the CD-FVD \cite{ge2024content} for spatial and temporal quality, the CLIP \cite{radford2021learning} for semantic consistency, the EgoVideo \cite{pei2024egovideo} for action consistency, the DOVER \cite{wu2023dover} for clarity score, frame interpolation model \cite{li2023amt} for motion smoothness, and RAFT \cite{teed2020raft} for motion strength. Additionally, following \cite{xu2024camco,he2024cameractrl}, we assess kinematic control consistency using translation error and rotation error, which measures the difference between COLMAP poses and the ground truth poses in the canonical space \cite{xu2024camco}. The specific calculations for each metric are detailed in the supplement.

\begin{figure*}[t]
\centering
\resizebox{1\linewidth}{!}{
\includegraphics{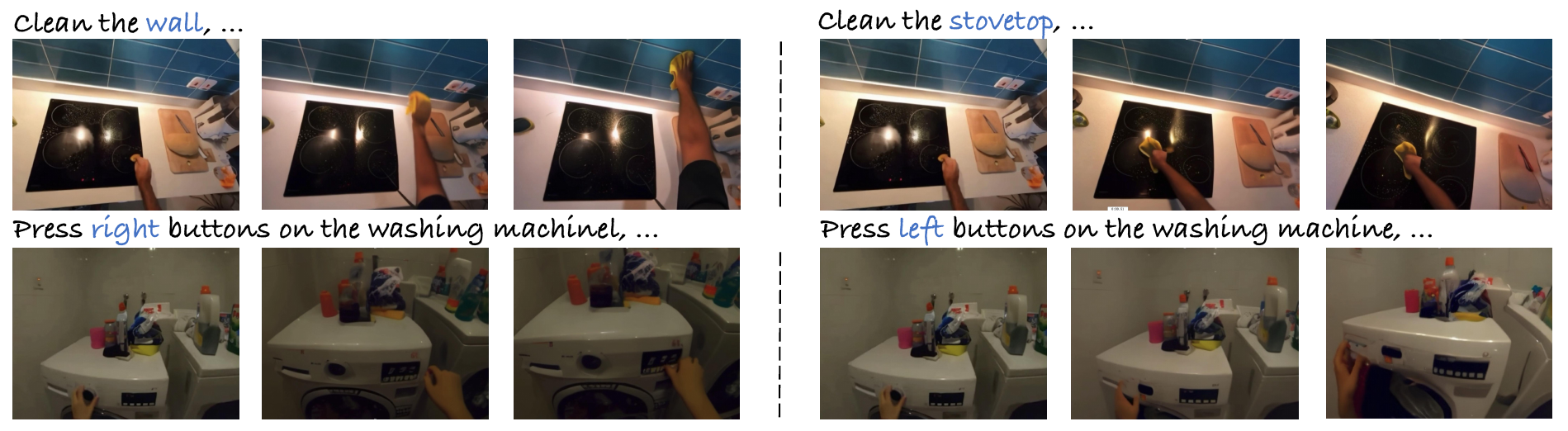}}
\caption{Visualizations show that \textit{EgoDreamer} can realize distinct action controls based on different text descriptions.}
\label{fig:egodreamer_text}
\vspace{-0.5em}
\end{figure*}

\begin{figure*}[h]
\centering
\resizebox{1\linewidth}{!}{
\includegraphics{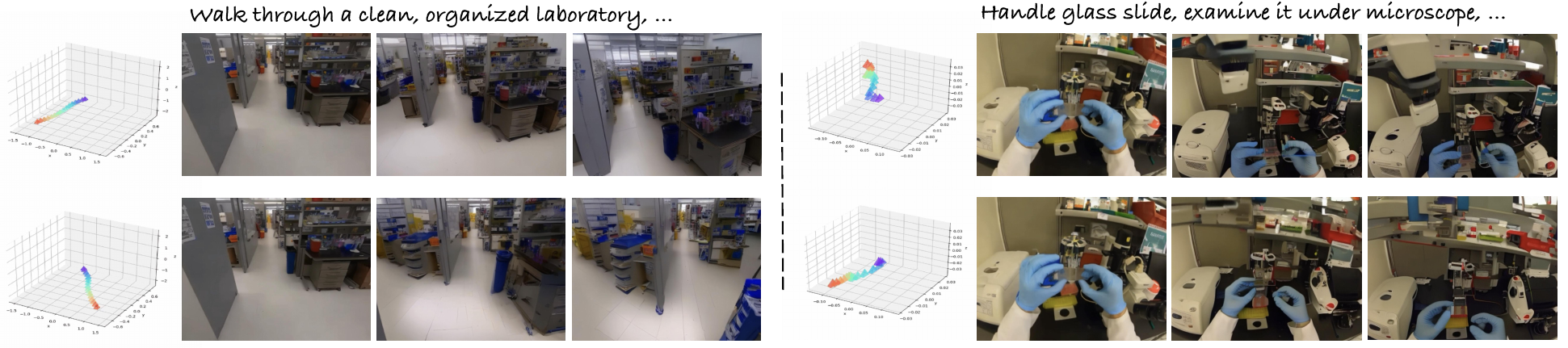}}
\caption{Visualizations demonstrate that \textit{EgoDreamer} can generate various egocentric videos based on different low-level commands.}
\label{fig:egodreamer_pose}
\vspace{-1em}
\end{figure*}

\begin{table*}[h]
    \centering
    \small
  \resizebox{0.95\linewidth}{!}{
  \begin{tabular}{ccccc|ccccc}
    \toprule
    \textit{w. EgoVid} &  ControlNet &  ControlNeXt & AA &  UAE  & CD-FVD $\downarrow$& Semantic Consistency $\uparrow$ & Action Consistency $\uparrow$ & Rot Err $\downarrow$ & Trans Err $\downarrow$ \\
    \midrule
      & \Checkmark & & &&241.90&0.263&0.490&5.32&9.27 \\
     \Checkmark  & \Checkmark &&&&238.87&0.266&0.493&4.01&8.66 \\
      \Checkmark  & \Checkmark &&&\Checkmark&239.01&0.268&0.494&3.58&8.41 \\
      \Checkmark  &  &\Checkmark&&\Checkmark&234.13&\textbf{0.269}&0.497&3.59&7.93\\
      
      \Checkmark  &  & &\Checkmark&\Checkmark&\textbf{229.82}&0.268&\textbf{0.498}&\textbf{3.28}&\textbf{7.62} \\
    \bottomrule
    \end{tabular}}
    \caption{Ablation study on training strategy and different components of \textit{EgoDreamer}.} 
    \label{tab:exp_egodreamer}
\end{table*}

Next, we verify the impact of different data cleaning strategies on egocentric video generation. Subsequently, we substantiate, quantitatively and qualitatively, that the proposed \textit{EgoVid} can enhance various baselines' egocentric video generation capabilities. Finally, experiments
are conducted to demonstrate that the proposed \textit{EgoDreamer} can generate egocentric videos under the control of both action descriptions and kinematic signals.

\subsection{Data Cleaning Strategy Comparison}
\label{sec:exp_filter}
In this subsection, we employ the state-of-the-art video diffusion model DynamiCrafter \cite{xing2023dynamicrafter} as the baseline, which is trained on the \textit{Image+Text-to-Video} task to evaluate various data cleaning strategies. 

\noindent
\textbf{Strategy-1.} This strategy focuses on ensuring text-frame consistency (with $\text{CLIP}_{TF} \ge 0.275$) and frame-frame consistency ($\text{CLIP}_{FF} \ge 0.8$). Additionally, we retained videos with an average optical flow $\ge$ 3 and a DOVER score $\ge$ 0.3. This process yielded a subset \textit{EgoVid-1M-1}. DynamiCrafter is finetuned for one epoch using this subset. As illustrated in Fig.~\ref{fig:clean}, this model achieved the highest semantic consistency metrics. However, the stringent criteria for both text-frame and frame-frame consistency favored the retention of videos with slow motion. Consequently, the motion strength of the generated videos falls below the baseline, which is not desirable for effective video generation.

\noindent
\textbf{Strategy-2.} The thresholds for text-frame consistency and frame-frame consistency are relaxed ($\text{CLIP}_{TF} \ge 0.27$, $\text{CLIP}_{FF} \ge 0.75$). Besides, we retain videos with an average optical flow between 3 and 40, and those with a DOVER score  $\ge$  0.3. This strategy results in a subset \textit{EgoVid-1M-2}. Upon finetuning DynamiCrafter for one full epoch, as shown in Fig.~\ref{fig:clean}, we observe a significant improvement in the motion strength. However, the accelerated motion introduces artifacts, leading to visual fragmentation. Consequently, this negatively impacts the text-frame semantic consistency, resulting in scores below the baseline.

\noindent
\textbf{Strategy-3.} we further relax the thresholds for text-frame consistency ($\text{CLIP}_{TF} \ge 0.26$) and frame-frame consistency ($\text{CLIP}_{FF} \ge 0.7$), while introducing an action consistency constraint (EgoVideo score $\ge$ 0.22). Videos are retained with an average optical flow between 3 and 35, as well as those with a DOVER score $\ge$ 0.3. Notably, as mentioned in Sec.~\ref{sec:clean}, we also retain videos with average optical flow values below 3, provided that the proportion of optical flow ($\ge$ 12 pixels) is greater than 3\%. This resulted in the \textit{EgoVid-1M-3} subset. Compared to the previous two strategies, the model finetuned on \textit{EgoVid-1M-3} effectively enhances both semantic and action consistency while ensuring moderate motion strength, achieving the best CD-FVD score. Furthermore, the \textit{5-point} optical flow filtering method allowed for a focus on local motion scenarios. As illustrated in Fig.~\ref{fig:clean_cmp}, strategy-3 accurately models intricate hand movements, in contrast to the stationary visuals of strategy-1 and the exaggerated motion of strategy-2.

\subsection{Enhancement in Egocentric Video Generation}
In this subsection, experiments are conducted to verify that the proposed \textit{EgoVid} enhances the egocentric video generation capabilities of various baselines. Specifically, SVD \cite{svd}, DynamiCrafter \cite{xing2023dynamicrafter}, and OpenSora \cite{opensora} are selected as baselines, which are initialized with their original weights, and then we employ \textit{EgoVid-1M-3} for finetuning. For training efficiency and fair comparison, we resize all input video to 480p and focus exclusively on the \textit{Image+Text-to-Video} tasks. As shown in Tab.~\ref{tab:exp_cmp}, the experiment results demonstrate that training with \textit{EgoVid} improves performance across all three baselines on six different metrics. Specifically, the \textit{EgoVid} finetuning significantly enhances the motion strength of egocentric videos while also improving consistency in text-video alignment, action-video alignment, and overall image clarity. Consequently, the CD-FVD metric shows a notable improvement. Additionally, we conduct a visualization comparison of different baselines before and after finetuning, as illustrated in Fig.~\ref{fig:vis_cmp}. Prior to \textit{EgoVid} finetuning, various baselines exhibit issues such as frame fragmentation and distortion in egocentric scenarios (e.g., appearance of incongruous objects and hand fragmentation). This underscores the inadequacy of most existing video generation models in egocentric contexts. However, after the \textit{EgoVid} finetuning, the generated videos not only achieve superior alignment with text prompts, but also exhibit enhancement in visual quality.

\subsection{EgoDreamer Experiments}

In this subsection, we conduct experiments to demonstrate that \textit{EgoDreamer} can generate egocentric videos under the control of both action descriptions and kinematic signals. Additionally, the efficacy of the proposed UAE and AA modules will be validated. In our experiments, we initialize \textit{EgoDreamer} using weights from \cite{xing2023dynamicrafter}. The results are presented in Tab.~\ref{tab:exp_egodreamer}. In Row-1, the low-level kinematic control signals are integrated via ControlNet \cite{ge2024content}, which resembles \cite{xu2024camco,he2024cameractrl}. Row-2 utilizes \textit{EgoVid-1M-3} to pre-train the model. Compared with Row-1, results indicate significant improvements across five metrics after \textit{EgoVid-1M-3} finetuning. In Row-3, we further introduce the UAE module to strengthen the association between low-level kinematic control and high-level action descriptions. The experimental results indicate that this enhancement further improves action alignment and reduces the deviation in low-level kinematic control compared to Row-2. In Row-4 and Row-5, we replace the ControlNet with ControlNext \cite{peng2024controlnext} and the AA module. The results reveal that the AA module exhibits superior performance compared to both ControlNet and ControlNext, as it facilitates learnable parameterized alignment from a multi-scale perspective. Finally, we visualize videos generated by \textit{EgoDreamer}, as depicted in Fig.~\ref{fig:egodreamer_text}. Under initial frame conditions, varying the input text descriptions enables \textit{EgoDreamer} to realize distinct action controls. Furthermore, as illustrated in Fig.~\ref{fig:egodreamer_pose}, with the same initial frame, the model can generate videos that incorporate a composite of multiple low-level kinematic controls. Notably, \textit{EgoDreamer} to produce videos with meter-level movements (e.g., walking) and centimeter-level nuanced movements (e.g., intricate hand actions in a laboratory environment). Additional visualizations can be found in the supplement.

\section{Discussion and Conclusion}
In this paper, we present \textit{EgoVid-5M}, which is the first high-quality dataset meticulously curated for egocentric video generation, comprising 5 million video clips enriched with detailed action annotations. This dataset effectively addresses the challenges associated with the dynamic nature of egocentric perspectives, the intricate diversity of actions, and the complex variety of encountered scenes.
The implementation of a sophisticated data cleaning pipeline further ensures the dataset's integrity and usability, maintaining frame consistency, action coherence, and motion smoothness under egocentric conditions. Additionally, we propose \textit{EgoDreamer}, which showcases the ability to generate egocentric videos by simultaneously incorporating action descriptions and kinematic control signals, thereby enhancing the realism and applicability of generated content.
We hope that the proposed \textit{EgoVid-5M} dataset, along with the associated annotations and metadata, will serve as a valuable resource for the research community. We encourage researchers to leverage these innovations to propel further exploration and development in the realm of egocentric video generation, ultimately advancing applications in virtual reality, augmented reality, and gaming.

{\small
\bibliographystyle{ieee_fullname}
\bibliography{PaperForReview}
}
\clearpage

% \title{Supplementary Material for \\EgoVid-5M: A Large-Scale Video-Action Dataset \\
% for Egocentric Videos Generation}
% \maketitle
In the supplement materials, we first elaborate on the annotation and cleaning details of \textit{EgoVid}. Then we present additional training details of different baselines and the proposed \textit{EgoDreamer}. Subsequently, the evaluation details are elaborated. Finally, we present additional visualizations.

\begin{figure*}[t]
\centering
\resizebox{1\linewidth}{!}{
\includegraphics{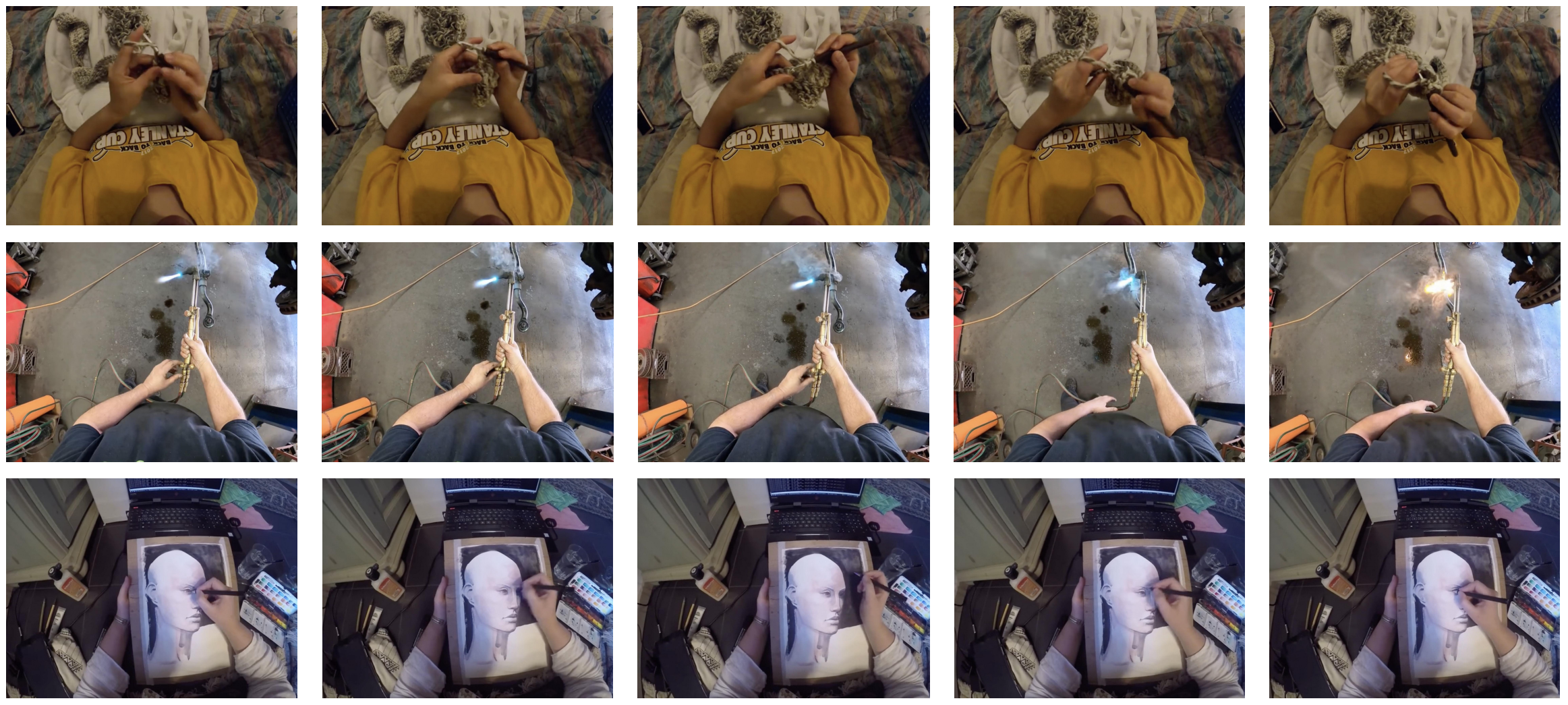}}
\caption{Videos cleaned from the \textit{five-point} optical flow strategy (average optical flow below 3, and the proportion of optical flow ($\ge$12 pixels) is greater than 3\%). This strategy retains videos with a static background while capturing detailed and extensive motion in hands.}
\label{fig:5pts-data}
\vspace{-1em}
\end{figure*}

\begin{figure*}[t]
\centering
\resizebox{1\linewidth}{!}{
\includegraphics{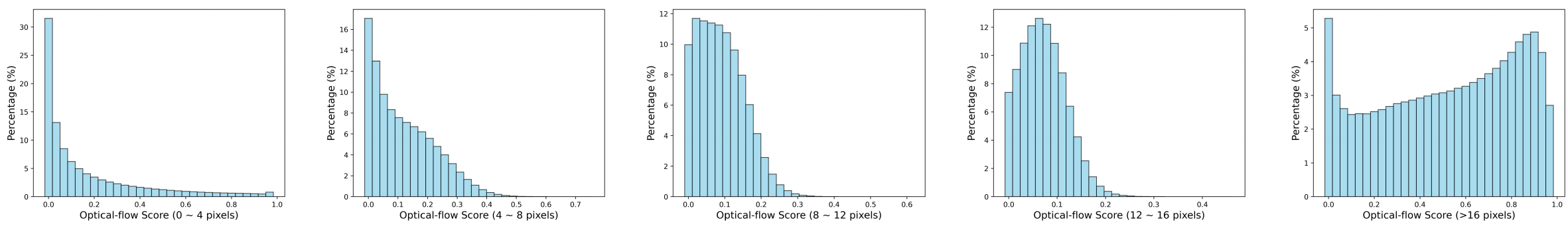}}
\caption{\textit{Five-point} optical flow distribution.}
\label{fig:5pts-dist}
\vspace{-1em}
\end{figure*}

\section{Annotation and Cleaning Details}

\subsection{Kinematic Annotation Details}

To enhance kinematic annotation accuracy, we fuse camera poses from IMU and ParticleSfM \cite{zhao2022particlesfm}, utilizing the Kalman filter. First, we filter the IMU data to remove gravitational components and noise. Next, we employ least squares estimation to determine the initial velocity and scale factor for the ParticleSfM poses. Finally, we align both the IMU poses and ParticleSfM results to the camera coordinate system (detailed explanations of these processes can be found in the main text). The Kalman filter implementation involves the following steps:

The state vector $\mathbf{x}=[x,y,z,q_1,q_2,q_3,q_4,v_x,v_y,v_z]$ is initialized from IMU pose to represent position, quaternion, and velocity. The error covariance matrix $\mathbf{P}$, process noise covariance $\mathbf{Q}$ and observation noise covariance F$\mathbf{R}$ are initialized as $0.1\cdot \mathbf{I}_{10 \times 10}$, $0.01\cdot \mathbf{I}_{10 \times 10}$ and $0.1\cdot \mathbf{I}_{7 \times 7}$. In the prediction step, the state transition function  $\mathbf{f}$ is applied to predict the next state:
\begin{equation}
    \mathbf{x}_{k \mid k-1}=\mathbf{f}\left(\mathbf{x}_{k-1}, \mathbf{u}_{k}\right),
\end{equation}
where $\mathbf{u}_{k}$ consists of IMU readings, and $\mathbf{f}$  predicts the next state by updating the current state through integration, incorporating the linear acceleration and angular velocity measured by the IMU. The covariance of the predicted state is updated as:
\begin{equation}
    \mathbf{P}_{k \mid k-1}=\mathbf{F P}_{k-1} \mathbf{F}^{T}+\mathbf{Q},
\end{equation}
where $\mathbf{F}$ is the Jacobian of the transition matrix. In the update phase, we compute the measurement residual $\mathbf{y}_k$:
\begin{equation}
    \mathbf{y}_k = \mathbf{x'}_k - \mathbf{Hx}_{k|k-1},
\end{equation}
where $\mathbf{x'}=[x',y',z',q'_1,q'_2,q'_3,q'_4]$ is the ParticleSfM pose, $H=\begin{bmatrix} 
\mathbf{I}_{3 \times 3} & \mathbf{0} \\ 
\mathbf{0} & \mathbf{I}_{4 \times 4} 
\end{bmatrix}
]$ is the Jacobian of the observation model.

The innovation covariance $\mathbf{S}_k$ is given by:
\begin{equation}
    \mathbf{S}_k=\mathbf{HP}_{k|k-1}\mathbf{H}^{T}+\mathbf{R},
\end{equation}
and the Kalman gain is calculated by:
\begin{equation}
    \mathbf{K}_k=\mathbf{P}_{k|k-1}\mathbf{H}^{T}\mathbf{S}_k^{-1}.
\end{equation}
The state estimate is then updated:
\begin{equation}
    \mathbf{x}_k=\mathbf{x}_{k|k-1} + \mathbf{K}_k\mathbf{y}_k.
\end{equation}
Finally, the error covariance matrix is updated:
\begin{equation}
    \mathbf{P}_k=(\mathbf{I}-\mathbf{K}_k\mathbf{H})\mathbf{P}_{k|k-1}.
\end{equation}
This iteration continues for each IMU reading, yielding a refined series of pose estimates.

\subsection{Data Cleaning Details}

\noindent
\textbf{\textit{Five-Point} Optical Flow Filtering.}
A typical approach to describe video motion strength is optical flow \cite{teed2020raft}. Therefore, we first represent video motion by averaging global optical flow. Notably, this approach only encapsulates the average motion magnitude. However, in egocentric scenarios, where a substantial portion of the scene remains static and only foreground elements (e.g., hands) exhibit motion, applying a filtering strategy based solely on average optical flow may result in the inadvertent exclusion of valuable, fine-grained hand movement data.
Therefore, as a supplement, we calculate the \textit{five-point} optical flow, which involves the proportion $P_{m\sim n}$ of optical flow score across different pixel intervals: 
\begin{equation}
    P_{m\sim n} = \frac{\sum_{x,y}\delta(m\le|F(x,y)|< n)}{N},
\end{equation}
where $N$ is the total pixel number, $F$ is the optical flow map, $\delta$ is the indicator function. Specifically, we calculate $P_{0\sim 4},P_{4\sim 8},P_{8\sim 12},P_{12\sim 16},P_{16\sim}$, their distribution is shown in Fig~\ref{fig:5pts-dist}. We performed data filtering based on the \textit{five-point} optical flow, as illustrated in Fig.~\ref{fig:5pts-data}, where the average optical flow magnitude is less than 3 pixels, and over 3\% of the pixels exhibit motion greater than 12 pixels. The figure shows that although most of the background elements remain static, the hand movements are dynamic and extensive. Such data are beneficial for training egocentric video generation with subtle hand motions.

\noindent
\textbf{Semantic Consistency Comparison.}
\begin{figure}[t]
\centering
\resizebox{1\linewidth}{!}{
\includegraphics{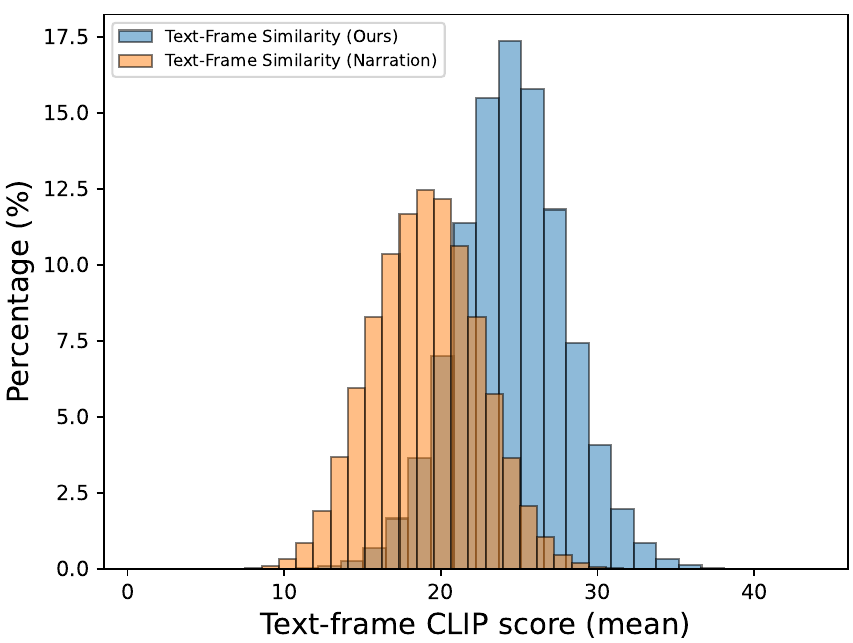}}
\caption{Semantic consistency comparison between our text annotation and the original human narration.}
\label{fig:clip-narr}
\vspace{-1em}
\end{figure}

In the Ego4D dataset, only human narrations are available as text annotations. However, these narrations are relatively simple and lack semantic alignment with the video frames. To address this, we first employ a multimodal large language model (MLLM) \cite{zhang2024llavanext-video} to generate detailed captions for the videos. Then, a large language model (LLM) \cite{qwen2} is used to summarize egocentric action descriptions from these detailed captions. We calculate the semantic consistency between captions and the frames using CLIP \cite{radford2021learning}. As shown in Fig.~\ref{fig:clip-narr}, the semantic similarity of our generated captions is significantly higher than that of the original human narrations.

\section{Training Details}
We validated the effectiveness of our \textit{EgoVid-5M} using video diffusion baselines with different architectures, including U-Net (SVD \cite{svd} and DynamiCrafter \cite{xing2023dynamicrafter}), and DiT (OpenSora \cite{opensora}). The training details are as follows: 
(1) For SVD, we employ the pre-trained 1.1 version\footnote{huggingface.co/stabilityai/stable-video-diffusion-img2vid-xt-1-1} and extend its \textit{img-to-video} architecture to an \textit{Image+Text-to-Video} setup. Specifically, we replace the image CLIP branch with a text CLIP branch\footnote{huggingface.co/openai/clip-vit-base-patch32}, which is aligned with the image CLIP version used in SVD. During training, input videos are resized to 480p, and we employed the EDM scheduler \cite{edm} with a learning rate of 1e-4 and a batch size of 64, finetuning on \textit{EgoVid-1M-3} for one epoch.
(2) For DynamiCrafter, we leverage the pre-trained model at 512 resolution \footnote{huggingface.co/Doubiiu/DynamiCrafter\_512}. Videos are resized to 480p during training, utilizing the DDPM scheduler \cite{ddpm} with a learning rate of 1e-5 and a batch size of 64. The finetuning was conducted on \textit{EgoVid-1M-3} for one epoch.
(3) For OpenSora, we used the pre-trained version 1.2 model\footnote{huggingface.co/hpcai-tech/OpenSora-STDiT-v3}, adjusting its data bucket strategy to train only on 480p inputs, and set mask ratios to mask only the first frame. The model was trained with the RF \cite{rf1,rf2} scheduler, a learning rate of 1e-4, and a batch size of 64, using \textit{EgoVid-1M-3} for one epoch.

\begin{figure*}[t]
\centering
\resizebox{1\linewidth}{!}{
\includegraphics{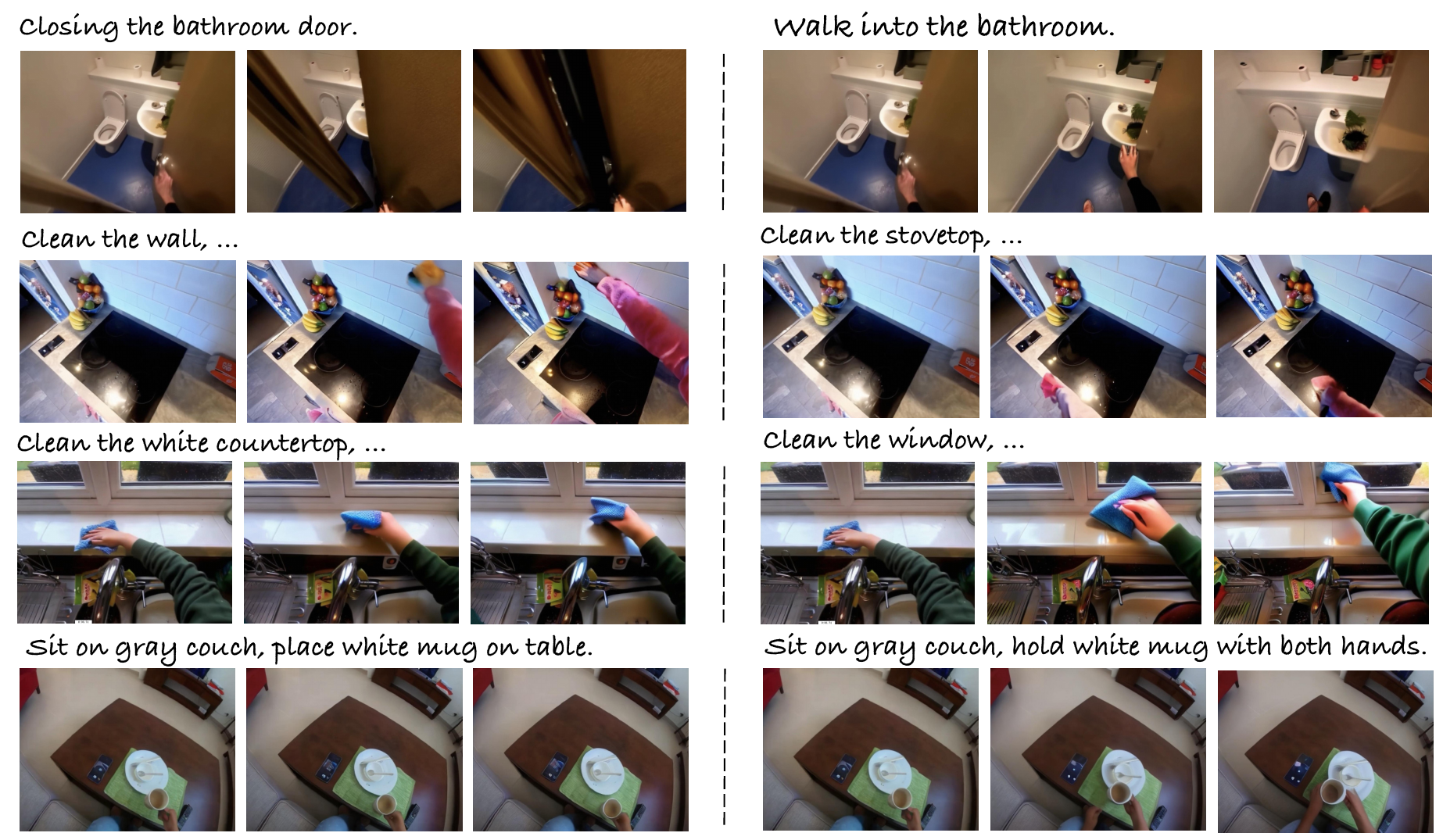}}
\caption{Visualizations showing that \textit{EgoDreamer} can generate action-driven egocentric videos based on high-level text descriptions.}
\label{fig:vis2}
\vspace{2em}
\end{figure*}

\begin{figure*}[h]
\centering
\resizebox{1\linewidth}{!}{
\includegraphics{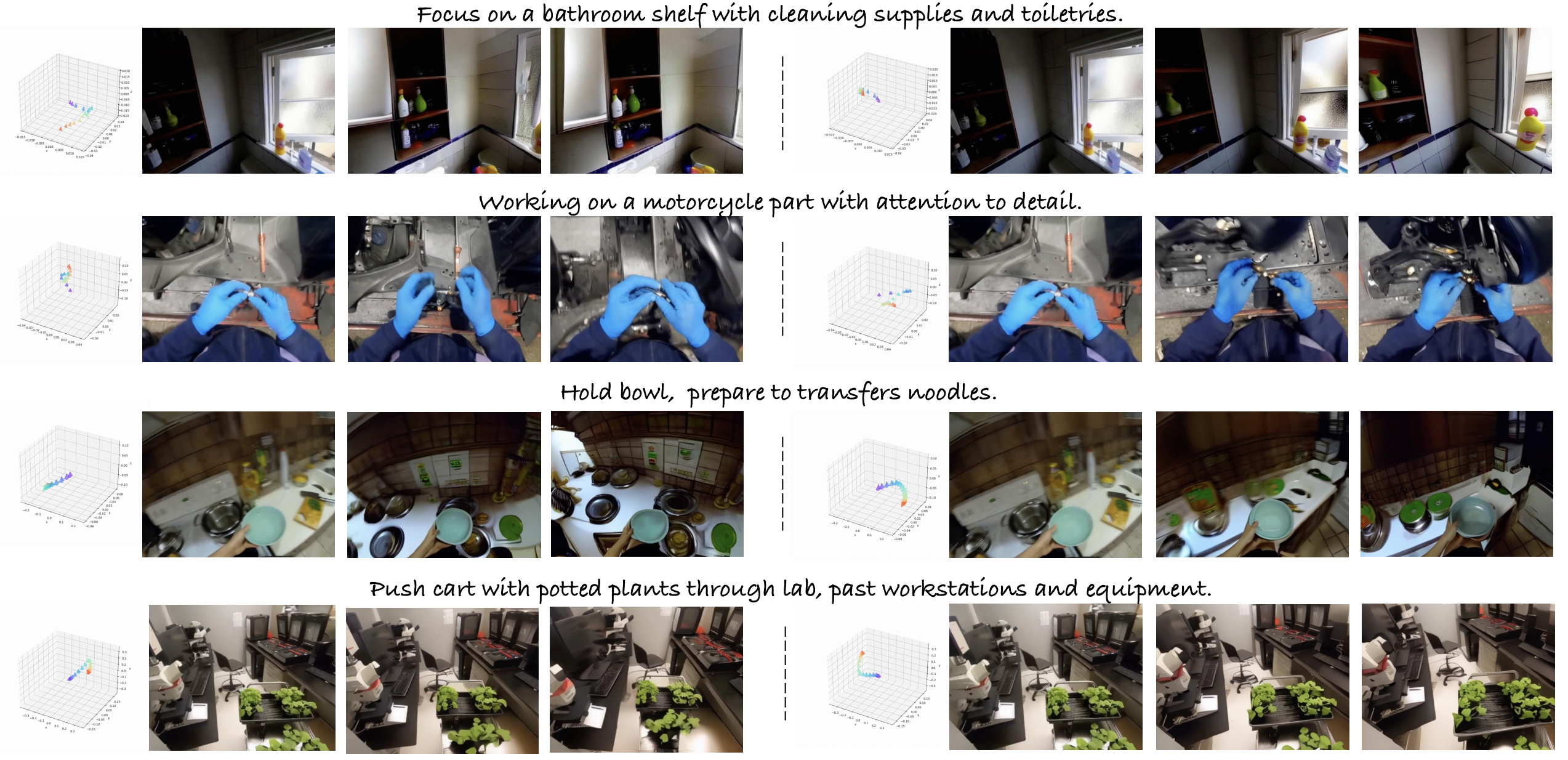}}
\caption{Visualizations showing that \textit{EgoDreamer} can generate action-driven egocentric videos based on low-level kinematic control.}
\label{fig:vis3}
\end{figure*}

\begin{figure*}[h]
\centering
\resizebox{1\linewidth}{!}{
\includegraphics{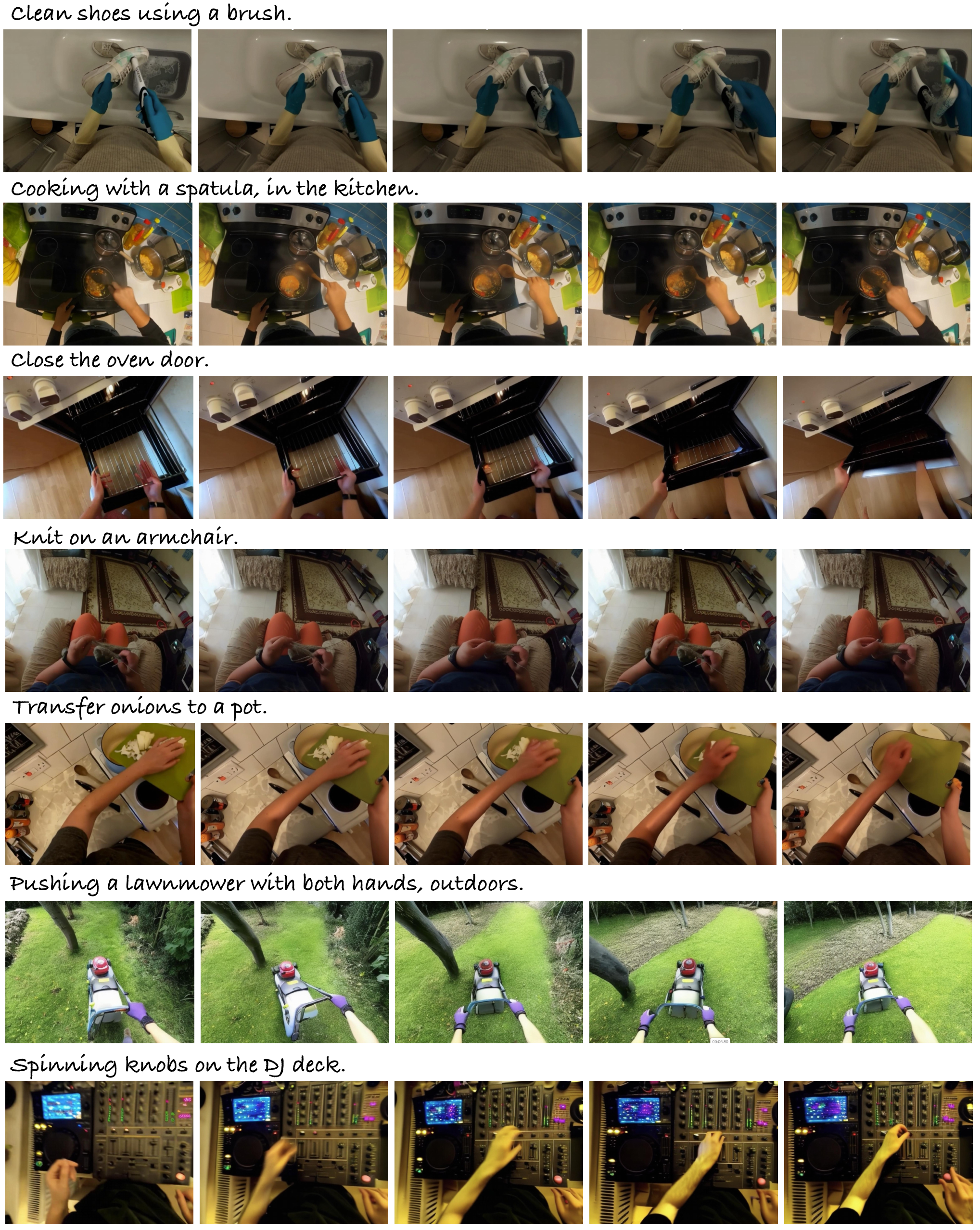}}
\caption{Visualizations verifying that \textit{EgoDreamer} can generate diverse egocentric videos based on action descriptions.}
\label{fig:vis1}
\end{figure*}

For \textit{EgoDreamer}, we first initialize it with the pre-trained model at 512 resolution \cite{xing2023dynamicrafter},
then \textit{EgoDreamer} are further trained on \textit{EgoVid-1M-3} to adapt to egocentric scenes, with batch size 64 and learning rate 1e-5. Finally, we finetune the proposed Unified Action Encoder (UAE) and Adaptive Alignment (AA) using \textit{EgoVid-65K}, with batch size 32 and learning rate 1e-5.

\section{Evaluation Details}
The evaluation metrics are mainly from AIGCBench \cite{fan2023aigcbench} and VBench \cite{huang2024vbench}, along with other metrics such as CD-FVD \cite{ge2024content}, EgoVideo score \cite{pei2024egovideo} and kinematic consistency (Translation Error and Rotation Error) \cite{he2024cameractrl,xu2024camco}. These metrics are as follows:

\noindent
\textbf{Overall Quality.} CD-FVD\footnote{github.com/songweige/content-debiased-fvd}  is utilized to measure spatial and temporal quality. Compared with traditional FVD \cite{fvd}, CD-FVD favors both quality and motion of video frames.

\noindent
\textbf{Semantic Consistency.} CLIP\footnote{huggingface.co/openai/clip-vit-large-patch14} \cite{radford2021learning}  is employed to calculate the semantic consistency of text and frames. We uniformly sample four frames from each generated video, calculate the similarity between each frame and the text using CLIP, and then compute the average similarity score.

\noindent
\textbf{Action Consistency.} EgoVideo\footnote{drive.google.com/file/d/1k6f1eRdcL17IvXtdX\_J8WxNbju2Ms3AW/view} \cite{pei2024egovideo}  is utilized to calculate the action consistency of text and frames. In this metric, four frames are uniformly sampled from each video to calculate the action similarity between frames and text.

\noindent
\textbf{Motion Strength.} We employed the optical flow score to quantify the motion strength in videos. Specifically, we utilized the RAFT model\footnote{github.com/princeton-vl/RAFT} \cite{teed2020raft} to calculate the optical flow score. For each video, we sampled frames at 8-frame intervals as input to the model. The motion strength of the video segment was then determined by averaging the optical flow scores across all sampled frames.

\noindent
\textbf{Motion Smoothness.} To assess the continuity of motion in the generated video, we utilize the AMT model\footnote{huggingface.co/lalala125/AMT/resolve/main/amt-s.pth} \cite{pei2024egovideo}. Specifically, for a generated video with frames $[f_0, f_1, ..., f_{2n-1}, f_{2n}]$, we remove the odd-numbered frames, resulting in $[f_0, f_2, ..., f_{2n}]$. The AMT model is then employed to interpolate the omitted frames $[\hat{f_1}, \hat{f_3}, ..., \hat{f}_{2n-1}]$. Finally, we compute the mean absolute error between the interpolated frames and the original ones.

\noindent
\textbf{Clarity.} We leverage DOVER\footnote{huggingface.co/teowu/DOVER/resolve/main/DOVER.pth} \cite{wu2023dover} to calculate the video clarity, and we use the fused score that focuses on both aesthetic perspective and technical perspective.

\noindent
\textbf{Kinematic Consistency.} Following \cite{xu2024camco,he2024cameractrl}, we assess kinematic consistency using translation error and rotation error, which measures the difference between COLMAP poses and the ground truth poses in the canonical space:
\begin{equation}
    \operatorname { RotErr }=\sum_{i=1}^{n} \arccos \frac{\operatorname{tr}\left(\mathbf{R}_\text{gen}^{i} \mathbf{R}_\text{gt}^{i \mathrm{~T}}\right)-1}{2},
\end{equation}
\begin{equation}
\operatorname{TransErr}=\sum_{i=1}^{n}\left\|\mathbf{T}_\text{gt}^{i}-\mathbf{T}_\text{gen}^{i}\right\|_{2},
\end{equation}
where $\mathbf{R}_\text{gen}^{i}, \mathbf{R}_\text{gt}^{i}$ are the generated and ground truth rotation matrix for the $i$-th frame. $\mathbf{T}_\text{gen}^{i}, \mathbf{T}_\text{gt}^{i}$ are translation vectors for the generated and ground truth camera translation in the $i$-th frame.

\section{Visualizations}
We conducted additional visualizations of the results generated by EgoDreamer. As shown in Fig.~\ref{fig:vis1}, \textit{EgoDreamer} can leverage action descriptions to generate diverse egocentric videos, encompassing scenes such as householding, cooking, knitting, gardening, and music. These videos include both subtle hand movements and more extensive movements involving walking. Furthermore, as illustrated in Fig.~\ref{fig:vis2}, given the same initial frame, changing the high-level text descriptions can generate egocentric videos that comply with semantic control. Lastly, as depicted in Fig.~\ref{fig:vis3}, given the same initial frame, altering the low-level kinematic control can generate egocentric videos that conform to pose control.

\end{document}